\pdfoutput=1

\documentclass[11pt]{article}

\usepackage[]{acl}

\usepackage{times}
\usepackage{latexsym}

\usepackage[T1]{fontenc}

\usepackage[utf8]{inputenc}

\usepackage{microtype}

\usepackage{tabularx} 
\usepackage{array}
\usepackage{multirow}
\usepackage{booktabs}
\usepackage{arydshln} 
\usepackage{makecell} 
\usepackage{graphicx}
\usepackage{url} 
\usepackage{amsthm,amsmath,amssymb} 
\usepackage{mathrsfs} 
\usepackage{bm} 
\usepackage{pifont} 
\usepackage{float}
\usepackage{lipsum}
\usepackage{soul}
\usepackage{caption}
\usepackage{subcaption}
\usepackage{enumitem}
\usepackage{bbm}
\usepackage{dsfont}

\newcommand*{\affaddr}[1]{#1} 
\newcommand*{\affmark}[1][*]{\textsuperscript{#1}}
\newcommand*{\email}[1]{\texttt{#1}}

\title{Global and Local Hierarchy-aware Contrastive Framework for Implicit Discourse Relation Recognition}

\author{
Yuxin Jiang\affmark[1,2]~~~~Linhan Zhang\affmark[3]~~~~Wei Wang\affmark[1,2,4]\\
\affaddr{\affmark[1]The Hong Kong University of Science and Technology (Guangzhou)}\\
\affaddr{\affmark[2]The Hong Kong University of Science and Technology}\\
\affaddr{\affmark[3]School of Computer Science and Engineering, The University of New South Wales}\\
\affaddr{\affmark[4]Guangzhou Municipal Key Laboratory of Materials Informatics, \\The Hong Kong University of Science and Technology (Guangzhou)}\\
\email{yjiangcm@connect.ust.hk},
\email{linhan.zhang@unsw.edu.au},
\email{weiwcs@ust.hk}
}

\begin{document}
\maketitle
\begin{abstract}
Due to the absence of explicit connectives, implicit discourse relation recognition (IDRR) remains a challenging task in discourse analysis.
The critical step for IDRR is to learn high-quality discourse relation representations between two arguments.
Recent methods tend to integrate the whole hierarchical information of senses into discourse relation representations for multi-level sense recognition. 
Nevertheless, they insufficiently incorporate the static hierarchical structure containing all senses (defined as \textit{global hierarchy}), and ignore the hierarchical sense label sequence corresponding to each instance (defined as \textit{local hierarchy}).
For the purpose of sufficiently exploiting global and local hierarchies of senses to learn better discourse relation representations, we propose a novel \textbf{G}l\textbf{O}bal and \textbf{L}ocal Hierarchy-aware Contrastive \textbf{F}ramework (GOLF), to model two kinds of hierarchies with the aid of \textit{multi-task learning} and \textit{contrastive learning}.
Experimental results on PDTB 2.0 and PDTB 3.0 datasets demonstrate that our method remarkably outperforms current state-of-the-art models at all hierarchical levels.
\footnote{Our code is publicly available at \url{https://github.com/YJiangcm/GOLF_for_IDRR}}
\end{abstract}

\section{Introduction}
Implicit discourse relation recognition (IDRR) aims to identify logical relations (named senses) between a pair of text segments (named arguments) without an explicit connective (e.g., \texttt{however}, \texttt{because}) in the raw text. 
As a fundamental task in discourse analysis, IDRR has benefitted a wide range of Natural Language Processing (NLP) applications such as question answering \cite{liakata2013discourse}, summarization \cite{cohan2018summarization}, information extraction \cite{tang2021event}, etc. 

\begin{figure}[!t]
	\centering 
	\includegraphics[width=\columnwidth]{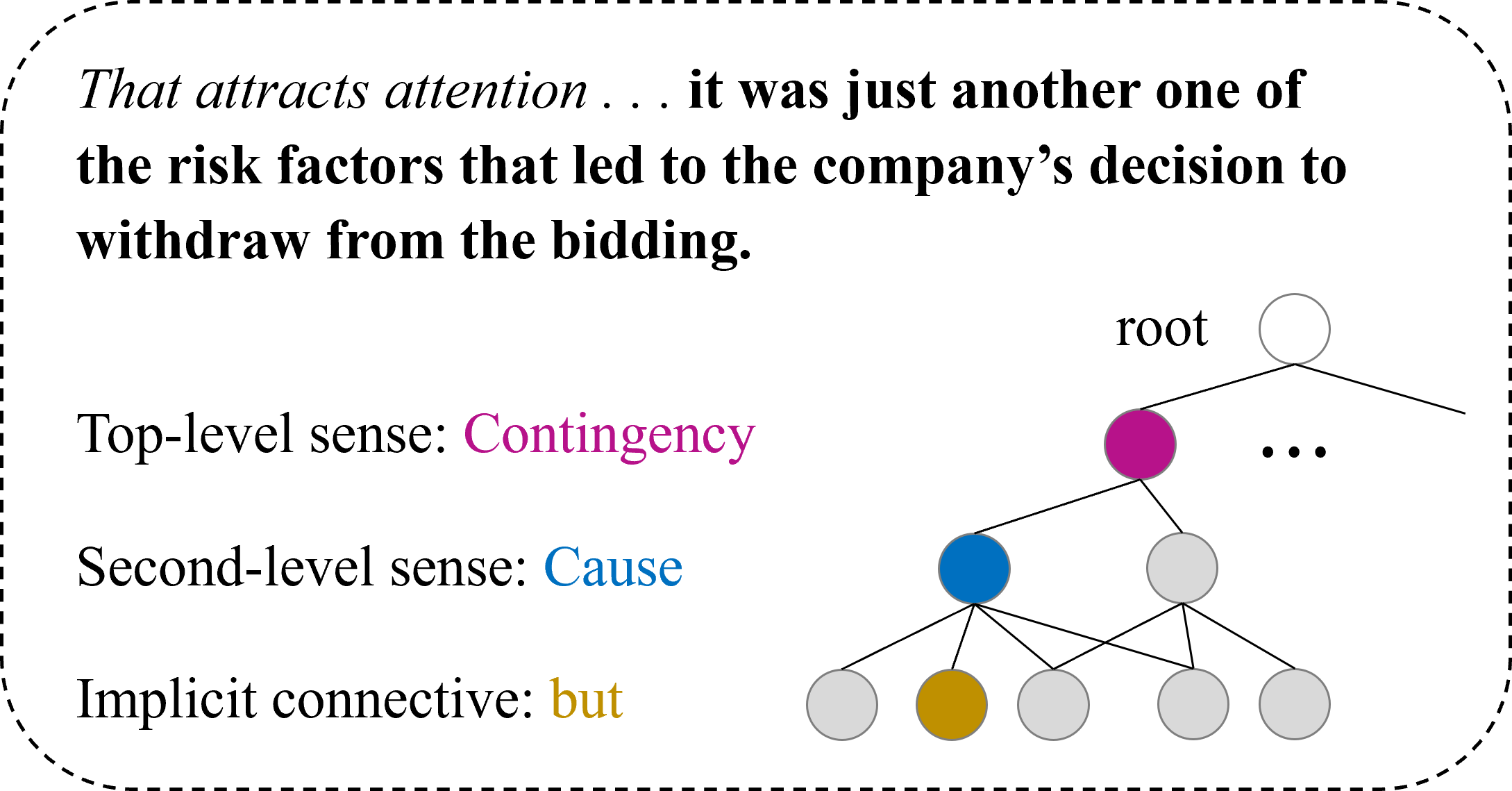}
	\caption{An IDRR instance in the PDTB 2.0 corpus \cite{prasad2008penn}. Argument 1 is in italics, and argument 2 is in bold. The implicit connective is not present in the original discourse context but is assigned by annotators. All senses defined in PDTB are organized in a three-layer hierarchical structure (defined as \emph{global hierarchy} in our paper), and the implicit connectives can be regarded as the most fine-grained senses.}
	\label{fig: example}
\end{figure}

The critical step for IDRR is to learn high-quality discourse relation representations between two arguments.
Early methods are dedicated to manually designing shallow linguistic features \cite{pitler2009automatic, park2012improving} or constructing dense representations relying on word embeddings \cite{liu2016nnma, dai2018pdrr, liu2020bmgf}. 
Despite their successes, they train multiple models to predict multi-level senses independently, while ignoring that the sense annotation of IDRR follows a hierarchical structure (as illustrated in Figure \ref{fig: example}).
To solve this issue, some researchers propose global hierarchy-aware models to exploit the prior probability of label dependencies based on Conditional Random Field (CRF) \cite{wu2020hiermtn} or the sequence generation model \cite{wu2022ldsgm}. 

\begin{figure}[!t]
	\centering 
	\includegraphics[width=\columnwidth]{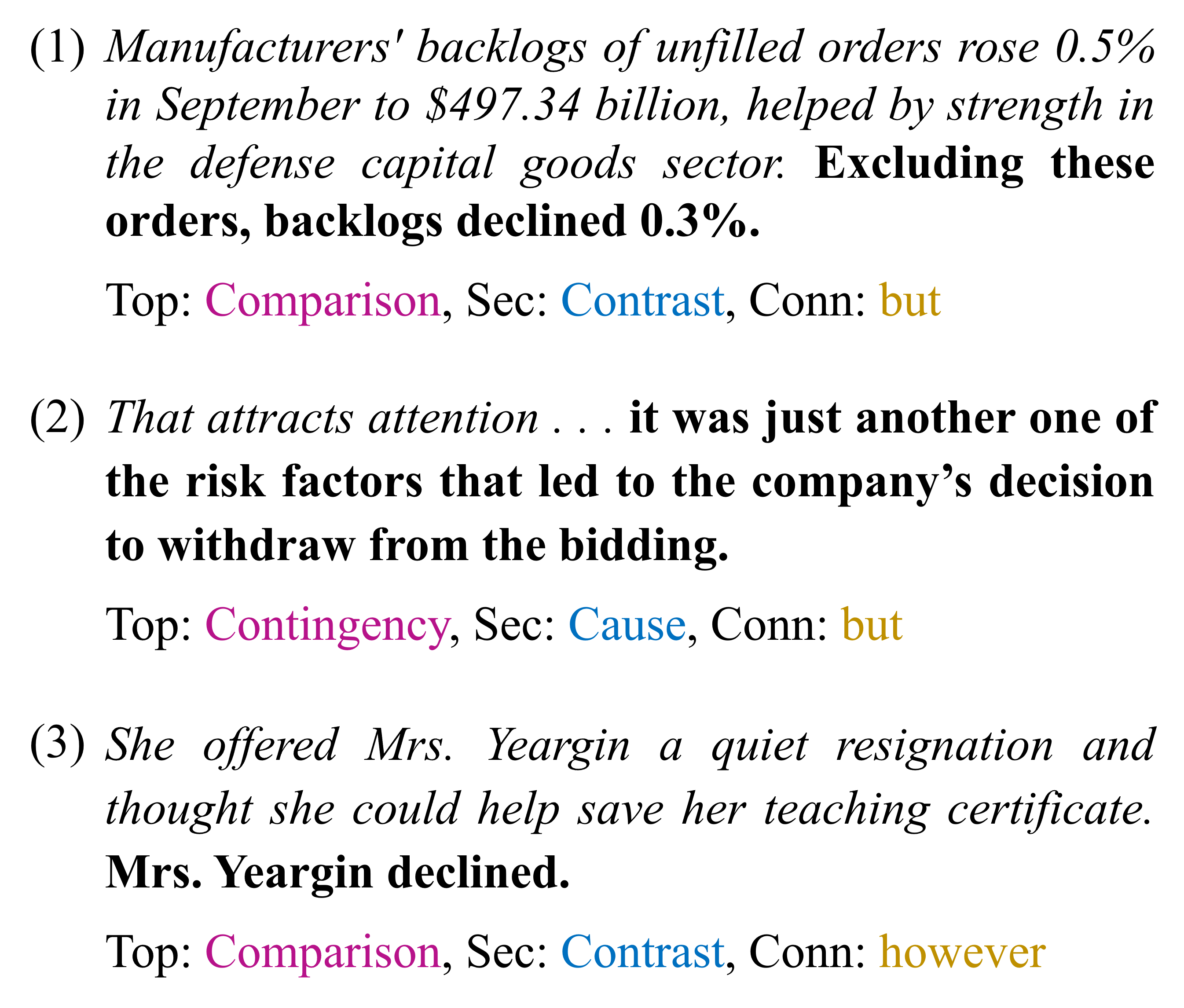}
	\caption{Three instances from PDTB 2.0. The sense label sequence of each instance is defined as \emph{local hierarchy} in our paper.}
	\label{fig: example1}
\end{figure}

However, existing hierarchy-aware methods still have two limitations.
\emph{Firstly}, 
though they exploit the fact that there are complex dependencies among senses and such information should be encoded into discourse relation representations, 
their manners of encoding the holistic hierarchical graph of senses may not be sufficient,
since they fail to strengthen the correlation between the discourse relation representation and its associated sense labels, which is highly useful for classification \cite{chen2020interaction}. 
\emph{Secondly}, they only consider the graph of the entire label hierarchy and ignore the benefit of the label sequence corresponding to each instance.
As shown in Figure \ref{fig: example1}, the label sequences of Instances (1) and (2) differ at both the top and second levels, while the label sequences of Instances (1) and (3) only differ at the most fine-grained level.
The similarity between label sequences provides
valuable information for regularizing discourse relation representations, e.g., by ensuring that the distance between representations of Instance (1) and (2) is farther than the distance between representations of Instance (1) and (3).
Under such an observation, we categorize the sense hierarchy into global and local hierarchies to fully utilize the hierarchical information in IDRR.
We define \emph{global hierarchy} as the entire hierarchical structure containing all senses, while \emph{local hierarchy} is defined as a hierarchical sense label sequence corresponding to each input instance.
Therefore, global hierarchy is static and irrelevant to input instances, while local hierarchy is dynamic and pertinent to input instances.

Built on these motivations, we raise our research question: \emph{How to sufficiently incorporate global and local hierarchies to learn better discourse relation representations?}
To this end, we propose a novel \textbf{G}l\textbf{O}bal and \textbf{L}ocal Hierarchy-aware Contrastive \textbf{F}ramework (GOLF), to inject additional information into the learned relation representation through additional tasks that are aware of the global and local hierarchies, respectively. This is achieved via the joint use of \emph{multi-task learning} and \emph{contrastive learning}.
The key idea of contrastive learning is to narrow the distance between two semantically similar representations, meanwhile, pushing away representations of dissimilar pairs \cite{Chen2020simclr, gao2021simcse}. It has achieved extraordinary successes in representation learning \cite{he2020momentum}. 
Finally, our multi-task learning framework consists of classification tasks and two additional contrastive learning tasks.
The global hierarchy-aware contrastive learning task explicitly matches textual semantics and label semantics in a text-label joint embedding space, which refines the discourse relation representations to be semantically similar to the target label representations while semantically far away from the incorrect label representations.
In the local hierarchy-aware contrastive learning task, we propose a novel scoring function to measure the similarity among sense label sequences.
Then the similarity is utilized to guide the distance between discourse relation representations.

The main contributions of this paper are three-fold:
\begin{itemize}[noitemsep]
\item We propose a novel global and local hierarchy-aware contrastive framework for IDRR, which sufficiently incorporates global and local hierarchies to learn better discourse relation representations.
\item To our best knowledge, our work is the first attempt to meticulously adapt contrastive learning to IDRR considering the global and local hierarchies of senses.
\item Comprehensive experiments and thorough analysis demonstrate that our approach delivers state-of-the-art performance on PDTB 2.0 and PDTB 3.0 datasets at all hierarchical levels, and more consistent predictions on multi-level senses.
\end{itemize}

\section{Related Work}
\subsection{Implicit Discourse Relation Recognition}
Early studies resort to manually-designed features to classify implicit discourse relations into four top-level senses \cite{pitler2009automatic, park2012improving}.
With the rapid development of deep learning, many methods explore the direction of building deep neural networks based on static word embeddings.
Typical works include shallow CNN \cite{zhang2015shallow}, LSTM with Multi-Level Attention \cite{liu2016nnma}, knowledge-augmented LSTM \cite{dai2018pdrr, dai2019regularization, guo2020kann}, etc.
These works aim to learn better semantic representations of arguments as well as capture the semantic interaction between them.
More recently, contextualized representations learned from large pre-trained language models (PLMs) and prompting \cite{schick2020s} have substantially improved the performance of IDRR.
More fined-grained levels of senses have been explored by \cite{liu2020bmgf, LongW22, chan2023discoprompt}.
Besides, researchers such as \cite{wu2020hiermtn, wu2022ldsgm} utilize the dependence between hierarchically structured sense labels to predict multi-level senses simultaneously.
However, these methods may be insufficient to exploit the global and local hierarchies for discourse relation representations.

\subsection{Contrastive Learning}
Contrastive learning is initially proposed in Computer Vision (CV) as a weak-supervised representation learning method, aiming to pull semantically close samples together and push apart dissimilar samples \cite{he2020momentum, Chen2020simclr}.
In NLP, contrastive learning has also achieved extraordinary successes in various tasks including semantic textual similarity (STS) \cite{gao2021simcse, shou2022amr, jiang2022promcse}, information retrieval (IR) \cite{hong2022sentence}, relation extraction (RE) \cite{chen2021cil}, etc.
Though intuitively supervised contrastive learning could be applied to IDRR through constructing positive pairs according to the annotated sense labels, it ignores the hierarchical structure of senses.
This paper is the first work to meticulously adapt contrastive learning to IDRR considering the global and local hierarchies of senses.

\section{Problem Definition}
Given $M$ hierarchical levels of defined senses $S=(S^1, ..., S^m, ..., S^M)$, where $S^m$ is the set of senses at the $m$-th hierarchical level, and a sample input consisting of two text spans, or $x_i=(arg_1, arg_2)$, our model aims to output a sequence of sense $y_i=(y_i^1, ..., y_i^m, ..., y_i^M)$, where $y_i^m \in S^m$.



\section{Methodology}

\begin{figure*}[!t]
	\centering 
	\includegraphics[width=\linewidth]{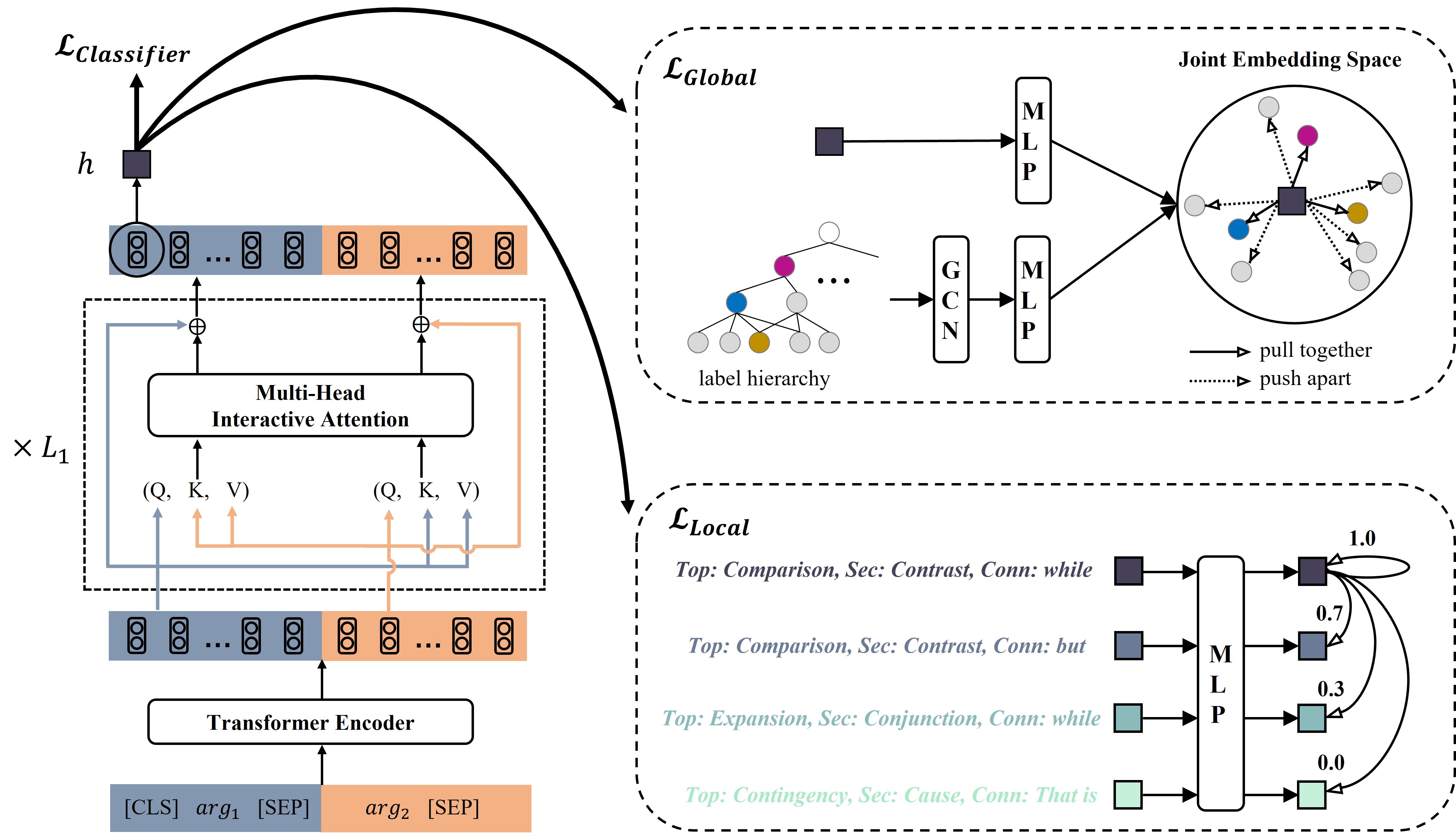}
	\caption{The overall architecture of our framework. The squares are denoted as discourse relation representations. Among the local hierarchy-aware contrastive loss $\mathcal{L}_{Local}$, we use colored squares to denote discourse relation representations of various instances in a mini-batch and list their sense label sequences on the left. Besides, note that the numbers on the right are similarity scores between sense label sequences calculated by our scoring function.}
	\label{fig: overview}
\end{figure*}

Figure \ref{fig: overview} illustrates the overall architecture of our multi-task learning framework.
Beginning at the left part of Figure \ref{fig: overview}, we utilize a Discourse Relation Encoder to capture the interaction between two input arguments and map them into a discourse relation representation $h$.
After that, the discourse relation representation $h$ is fed into a Staircase Classifier to perform classification at three hierarchical levels dependently.
While training, we will use two additional tasks, the global hierarchy-aware contrastive loss $\mathcal{L}_{Global}$ (in the upper right part of Figure \ref{fig: overview}) and the local hierarchy-aware contrastive loss $\mathcal{L}_{Local}$ (in the lower right part of Figure \ref{fig: overview}) as additional regularization to refine the discourse relation representation $h$.
During inference, we only use the Discourse Relation Encoder and the Staircase Classifier for classification and \emph{discard} the Global and Local Hierarchy-aware Contrastive Learning modules.
%
%
Detailed descriptions of our framework are given below.

\subsection{Discourse Relation Encoder}
\label{sec: Discourse Relation Encoder}
Given an instance $x_i=(arg_1, arg_2)$, we concatenate the two arguments and formulate them as a sequence with special tokens: $\verb![CLS]! \ arg_1 \ \verb![SEP]! \ arg_2 \ \verb![SEP]!$, where \texttt{[CLS]} and \texttt{[SEP]} denote the beginning and the end of sentences, respectively.
Then we feed the sequence through a Transformer \cite{vaswani2017transformer} encoder to acquire contextualized token representations $H$.
Previous works \cite{liu2016nnma, liu2020bmgf} indicate that deep interactions between two arguments play an important role in IDRR. 
To this end, we propose a Multi-Head Interactive Attention (MHIA) module to facilitate bilateral multi-perspective matching between $arg_1$ and $arg_2$.
As shown in the left part of Figure \ref{fig: overview}, we separate $H$ into $H_{arg_1}$ and $H_{arg_2}$, denoting as the contextualized representations of $arg_1$ and $arg_2$.
Then MHIA reuses the Multi-Head Attention (MHA) in Transformer, but the difference is that we take $H_{arg_1}$ as \textit{Query}, $H_{arg_2}$ as \textit{Key} and \textit{Value} and vice versa.
The intuition behind MHIA is to simulate human’s transposition thinking process: respectively considering each other’s focus from the standpoint of $arg_1$ and $arg_2$.
Note that the MHIA module may be stacked for $L_{1}$ layers. 
Finally, we use the representation of \texttt{[CLS]} in the last layer as the discourse relation representation and denote it as $h$ for simplicity.

\subsection{Staircase Classifier}
Given the discourse relation representation $h_i$ of an instance, we propose a "staircase" classifier inspired by \cite{abbe2021staircase} to output the label logits $t_{i}^m$ at each hierarchical level $m \in [1, M]$ in a top-down manner, where the higher-level logits are used to guide the logits at the current level:

\begin{small}
\begin{equation}
t_{i}^m = h_{i}W_{1}^m + t_{i}^{m-1}W_{2}^m + b^m
\end{equation}
\end{small}%
where $W_{1}^m \in \mathbb{R}^{d_{h} \times |S^m|}$, $W_{2}^m \in \mathbb{R}^{|S^{m-1}| \times |S^m|}$, $b^m \in \mathbb{R}^{|S^m|}$, $t_{i}^0=\vec{0}$.
Then the cross-entropy loss of the classifier is defined as follows:

\begin{small}
\begin{equation}
        \mathcal{L}_{CE} = -\frac{1}{|N|}\sum_{i \in N}\sum_{m=1}^M
\mathbb{E}_{\vec{y}_{i}^m}[\mathrm{LogSoftmax}(t_{i}^m)]
\end{equation}
\end{small}where $\vec{y}_{i}^m$ is the one-hot encoding of the ground-truth sense label $y_{i}^m$.

\subsection{Global Hierarchy-aware Contrastive Learning}
\label{sec: Global Hierarchy-aware Contrastive Learning}
The Global Hierarchy-aware Contrastive Learning module first exploits a Global Hierarchy Encoder to encode global hierarchy into sense label embeddings. Then, it matches the discourse relation representation of an input instance with its corresponding sense label embeddings in a joint embedding space based on contrastive learning.

\subsubsection{Global Hierarchy Encoder}
\label{sec: Sense Label Encoder}
To encode label hierarchy in a global view, we regard the hierarchical structure of senses as an undirected graph, where each sense corresponds to a graph node. 
Then we adopt a graph convolutional network (GCN) \cite{welling2016gcn} to induce node embeddings for each sense based on properties of their neighborhoods.
The adjacent matrix $A \in \mathbb{R}^{|S| \times |S|}$ is defined as follows:

\begin{small}
\begin{equation}
A_{ij} = \begin{cases}
1, & if \ i=j; \\
1, & if \ child(i)=j \ or \ child(j)=i; \\
0, & otherwise.
\end{cases}
\end{equation}
\end{small}where $S$ is the set of all senses, $i,j \in S$, $child(i)=j$ means that sense $j$ is the subclass of sense $i$. By setting the number layer of GCN as $L_2$, given the initial representation of sense $i$ as $r_{i}^{0} \in \mathbb{R}^{d_{r}}$, GCN updates the sense embeddings with the following layer-wise propagation rule:

\begin{small}
\begin{equation}
r_{i}^{l} = ReLU(\sum_{j \in S}D_{ii}^{-\frac{1}{2}}A_{ij}D_{jj}^{-\frac{1}{2}}r_{j}^{l-1}W^{l}+b^{l})
\end{equation}
\end{small}%
where $l \in [1, L_{2}]$, $W^{l} \in \mathbb{R}^{d_{r} \times d_{r}}$ and $b^{l} \in \mathbb{R}^{d_{r}}$ are learnable parameters at the $l$-th GCN layer, $D_{ii}=\sum_{j}A_{ij}$.
Finally, we take the output $\{r_{i}^{L_2}\}_{i \in S}$ of the $L_2$-th layer as the sense embeddings and denote them as $\{r_{i}\}_{i \in S}$ for simplicity.

\subsubsection{Semantic Match in a Joint Embedding Space}
\label{Semantic Match in a Joint Embedding Space}
In this part, we match textual semantics and label semantics in a text-label joint embedding space where correlations between text and labels are exploited, as depicted in the upper right part of Figure \ref{fig: overview}.
We first project the discourse relation representation $h_i$ of an instance $x_i$ and the sense label embeddings $\{r_{i}\}_{i \in S}$ into a common latent space by two different Multi-Layer Perception (MLP) $\Phi_1$ and $\Phi_2$.
Then, we apply a contrastive learning loss to capture text-label matching relationships, by regularizing the discourse relation representation to be semantically similar to the target label representations and semantically far away from the incorrect label representations:

\begin{small}
\begin{align}
\mathcal{L}_{G}&=-\frac{1}{|N|}
\sum_{i \in N} \sum_{j \in S}
\mathds{1}_{j \in y_{i}} \nonumber\\
&\times \log \frac{\exp\left(sim\Big(\Phi_1(h_i), \Phi_2(r_{j})\Big)/ \tau\right)}
{\sum_{j \in S} \exp\left(sim \Big(\Phi_1(h_i), \Phi_2(r_j)\Big)/ \tau\right)}
\end{align}
\end{small}where $N$ denotes a batch of training instances, $y_{i}$ is the sense label sequence of instance $x_i$,  $sim(\cdot)$ is the cosine similarity function, $\tau$ is a temperature hyperparameter. 
By minimizing the global hierarchy-aware contrastive learning loss, the distribution of discourse relation representations is refined to be similar to the label distribution.

Here we would like to highlight the key differences between our model and LDSGM \cite{wu2022ldsgm}, since we both utilize a GCN to acquire label representations.
Firstly, We use a different approach to capture the associations between the acquired label representations and the input text. In \cite{wu2022ldsgm}, the associations are \emph{implicitly} captured using the usual attention mechanism. In contrast, our model \emph{explicitly} learns them by refining the distribution of discourse relation representations to match the label distribution using contrastive learning.
Secondly, our work introduces a novel aspect that has been overlooked by earlier studies including \cite{wu2022ldsgm}: the utilization of local hierarchy information, which enables our model to better differentiate between similar discourse relations and achieve further improvements.

\subsection{Local Hierarchy-aware Contrastive Learning}
Following \cite{gao2021simcse}, we duplicate a batch of training instances $N$ as $N^+$ and feed $N$ as well as $N^+$ through our Discourse Relation Encoder E with diverse dropout augmentations to obtain $2|N|$ discourse relation representations. Then we apply an MLP layer $\Phi_3$ over the representations, which is shown to be beneficial for contrastive learning \cite{Chen2020simclr}.

To incorporate local hierarchy into discourse relation representations, it is tempting to directly apply supervised contrastive learning \cite{gunel2021scl} which requires positive pairs to have identical senses at each hierarchical level $m \in [1, M]$:

\begin{small}
\begin{align}
\mathcal{L}_{L'}&=-\frac{1}{|N|}\sum_{i \in N}\sum_{j \in N^+}
\left(\prod_{m=1}^M \mathds{1}_{y_{i}^{m} = y_{j}^{m}} \right) \nonumber\\
&\times \log \frac{\exp\left(sim\Big(\Phi_3(h_i), \Phi_3(h_j) \Big)/ \tau\right)}
{\sum_{j \in N^+} \exp\left(sim\Big(\Phi_3(h_i), \Phi_3(h_j) \Big)/ \tau\right)}
\label{equation: hard local}
\end{align}
\end{small}However, Equation (\ref{equation: hard local}) ignores the more subtle semantic structures of the local hierarchy, since it only admits positive examples as having \emph{identical}, no account for examples with highly similar annotations.
To illustrate, consider Instances (1) and (3) in Figure \ref{fig: example1}, where their sense label sequences only differ at the most fine-grained level. 
However, they are regarded as a negative pair in Equation (\ref{equation: hard local}), rather than a "relatively" positive pair.
The standard of selecting positive pairs is too strict in Equation (\ref{equation: hard local}), thus may result in semantically similar representations being pulled away.
To loosen this restriction, we regard all instance pairs as positive pairs but assign the degree of positive, by using a novel scoring function to calculate the similarity among label sequences $y_i=(y_i^1, ..., y_i^m, ..., y_i^M)$ and $y_j=(y_j^1, ..., y_j^m, ..., y_j^M)$.

In our case, there exist three hierarchical levels including Top, Second, and Connective, and we use $\mathbb{T}$, $\mathbb{S}$, and $\mathbb{C}$ to denote them. Consequently, there are in total $K=6$ sub-paths in the hierarchies, i.e., $P=\{\mathbb{T}, \mathbb{S}, \mathbb{C}, \mathbb{TS}, \mathbb{SC}, \mathbb{TSC}\}$.
Then we calculate the Dice similarity coefficient for each sub-path among the hierarchical levels and take the average as the similarity score between $y_i$ and $y_j$, which is formulated below:

\begin{small}
\begin{gather}
Score(y_i, y_j) = \frac{1}{K}\sum_{k=1}^K Dice(P_{i}^k, P_{j}^k)
\label{equation: score}
\end{gather}
\end{small}where $Dice(A,B)=(2|A \cap B|)/(|A| + |B|)$, $P_{i}^k$ is the $k$-th sub-path label set of $y_i$. 
Taking Instances (1) and (3) in Figure \ref{fig: example1} as examples, their label sequences are \emph{Top: Comparison, Sec: Contrast, Conn: but} and \emph{Top: Comparison, Sec: Contrast, Conn: however}, respectively.
Then the similarity score would be $\frac{1}{6}(\frac{2\times 1}{1+1} + \frac{2\times 1}{1+1} + \frac{2\times 0}{1+1} + \frac{2\times 2}{2+2} + \frac{2\times 1}{2+2} + \frac{2\times 2}{3+3})\approx0.7$.

Finally, our local hierarchy-aware contrastive loss utilizes the similarity scores to guide the distance between discourse relation representations:

\begin{small}
\begin{align}
\mathcal{L}_{L}&=-\frac{1}{|N|}\sum_{i \in N}\sum_{j \in N^+}
Score(y_i, y_j) \nonumber\\ 
&\times \log \frac{\exp\left(sim\Big(\Phi_3(h_i), \Phi_3(h_j) \Big)/ \tau\right)}
{\sum_{j \in N^+} \exp\left(sim\Big(\Phi_3(h_i), \Phi_3(h_j) \Big)/ \tau\right)}
\label{equation: soft local}
\end{align}
\end{small}%
Compared with Equation (\ref{equation: hard local}), Equation (\ref{equation: soft local}) considers more subtle semantic structures of the local hierarchy for selecting positive pairs.
It increases the relevance of representations for all similarly labeled instances and only pushes away instances with entirely different local hierarchies.
Thus, the local hierarchical information is sufficiently incorporated into discourse relation representations.

The overall training goal is the combination of the classification loss, the global hierarchy-aware contrastive loss, and the local hierarchy-aware contrastive loss:

\begin{small}
\begin{align}
\mathcal{L} = \mathcal{L}_{CE}
+ \lambda_1 \cdot \mathcal{L}_{G}
+ \lambda_2 \cdot \mathcal{L}_{L}
\end{align}
\end{small}where $\lambda_1$ and $\lambda_2$ are coefficients for the global and local hierarchy-aware contrastive loss, respectively.
We set them as 0.1 and 1.0 while training, according to hyperparameter search (in Appendix \ref{sec: appendix b}).

\section{Experiments}

\subsection{Dataset}

\paragraph{The Penn Discourse Treebank 2.0 (PDTB 2.0)}
PDTB 2.0 \cite{prasad2008penn} is a large-scale English corpus annotated with information on discourse structure and semantics. PDTB 2.0 has three levels of senses, i.e., classes, types, and sub-types. Since only part of PDTB instances is annotated with third-level senses, we take the top-level and second-level senses into consideration and regard the implicit connectives as third-level senses. There are 4 top-level senses including Temporal (Temp), Contingency (Cont), Comparison (Comp), and Expansion (Expa). Further, there exist 16 second-level senses, but we only consider 11 major second-level implicit types following previous works \cite{liu2020bmgf, wu2022ldsgm}. For the connective classification, we consider all 102 connectives defined in PDTB 2.0. 

\paragraph{The Penn Discourse Treebank 3.0 (PDTB 3.0)}
PDTB 3.0 \cite{webber2019pdtb3} is the updated version of PDTB 2.0, which includes an additional 13K annotations and corrects some inconsistencies in PDTB 2.0. 
Following the preprocess of PDTB 2.0, we consider 4 top-level senses, 14 majority second-level senses, and all 186 connectives defined in PDTB 3.0.

Appendix \ref{sec: appendix a} shows the detailed statistics of the PDTB corpora.
We follow early works \cite{ji2015pdtb, liu2020bmgf, wu2022ldsgm} using Sections 2-20 of the corpus for training, Sections 0-1 for validation, and Sections 21-22 for testing.
In PDTB 2.0 and PDTB 3.0, there are around 1\% data samples with multiple annotated senses. 
Following \cite{qin2016shallow}, we treat them as separate instances during training for avoiding ambiguity. At test time, a prediction matching one of the gold types is regarded as the correct answer.

\subsection{Baselines}
To validate the effectiveness of our method, we contrast it with the most advanced techniques currently available. As past research generally assessed one dataset (either PDTB 2.0 or PDTB 3.0), we utilize distinct baselines for each. Due to PDTB 3.0's recent release in 2019, there are fewer baselines available for it compared to PDTB 2.0.

\paragraph{Baselines for PDTB 2.0}
\begin{itemize}[]
\item  \textbf{NNMA} \cite{liu2016nnma}: a neural network with multiple levels of attention.
\item  \textbf{KANN} \cite{guo2020kann}: a knowledge-enhanced attentive neural network.
\item  \textbf{PDRR} \cite{dai2018pdrr}: a paragraph-level neural network that models inter-dependencies between discourse units as well as discourse relation continuity and patterns.
\item  \textbf{IDRR-Con} \cite{shi2019idrrcon}: a neural model that leverages the inserted connectives to learn
better argument representations.
\item  \textbf{IDRR-C\&E} \cite{dai2019regularization}: a neural model leveraging external event knowledge and coreference relations.
\item  \textbf{MTL-MLoss} \cite{nguyen2019mtlmlloss}: a neural model which predicts the labels and connectives simultaneously.
\item  \textbf{HierMTN-CRF} \cite{wu2020hiermtn}: a hierarchical multi-task neural network with a conditional random field layer.
\item  \textbf{BERT-FT} \cite{Kishimoto2020bertft}: a model applying three additional training tasks.
\item  \textbf{RoBERTa (Fine-tuning)}: a RoBERTa-based model fine-tuned on three sense levels separately.
\item  \textbf{BMGF-RoBERTa} \cite{liu2020bmgf}: a RoBERTa-based model with bilateral multi-perspective matching and global information fusion.
\item  \textbf{LDSGM} \cite{wu2022ldsgm}: a label dependence-aware sequence generation model.
\item  \textbf{ChatGPT} \cite{chatgpt23}: a ChatGPT-based method equipped with an in-context learning prompt template.
\end{itemize}

\paragraph{Baselines for PDTB 3.0}
\begin{itemize}[]
\item  \textbf{MANF} \cite{xiang2022manf}:  a
multi-attentive neural fusion model to encode and fuse both semantic connection and linguistic evidence.
\item  \textbf{RoBERTa (Fine-tuning)}: a RoBERTa-based model fine-tuned on three sense levels separately.
\item  \textbf{BMGF-RoBERTa} \cite{liu2020bmgf}: we reproduce the model on PDTB 3.0.
\item  \textbf{LDSGM} \cite{wu2022ldsgm}: we reproduce the model on PDTB 3.0.
\item  \textbf{ConnPrompt} \cite{xiang2022connprompt}: a PLM-based model using a connective-cloze Prompt to transform the IDRR task as a connective-cloze prediction task.
\end{itemize}

\renewcommand{\dblfloatpagefraction}{.9}
\begin{table*}[!tbp]
\small
\centering
\begin{tabular}{llcccccc}
\toprule
\multirow{2}{*}{\textbf{Model}} & \multirow{2}{*}{\textbf{Embedding}} & \multicolumn{2}{c}{\textbf{Top-level}} & \multicolumn{2}{c}{\textbf{Second-level}} & \multicolumn{2}{c}{\textbf{Connective}} \\
                       &                             & $F_{1}$            & $Acc$           & $F_{1}$              & $Acc$            & $F_{1}$            & $Acc$            \\
\midrule
\midrule
\multicolumn{8}{c}{\textit{PDTB 2.0}}
\\\midrule
NNMA \cite{liu2016nnma} & GloVe & 46.29 & 57.57 & - & - & - & - \\
KANN \cite{guo2020kann}                   & GloVe                       & 47.90              & 57.25              & -                & -               & -             & -              \\
PDRR \cite{dai2018pdrr} & word2vec & 48.82 & 57.44 & - & - & - & - \\
IDRR-Con \cite{shi2019idrrcon}               & word2vec                    & 46.40              & 61.42              & -                & 47.83               & -             & -              \\
IDRR-C\&E \cite{dai2019regularization}            & ELMo                        &  52.89             & 59.66              & 33.41                & 48.23               & -             & -              \\
MTL-MLoss \cite{nguyen2019mtlmlloss}             & ELMo                        & 53.00              & -              & -                & 49.95               & -             & -              \\
HierMTN-CRF \cite{wu2020hiermtn}           & BERT                        & 55.72              & 65.26              & 33.91                & 53.34               & 10.37              & 30.00               \\
BERT-FT \cite{Kishimoto2020bertft}               & BERT                        & 58.48              & 65.26              & -                & 54.32               & -             & -              \\
RoBERTa (Fine-tuning)             & RoBERTa                        & 62.96              & 69.98              & 40.34                & 59.87               & 10.06             & 31.45              \\
BMGF-RoBERTa \cite{liu2020bmgf}          & RoBERTa                     & 63.39              & 69.06              & -                & 58.13               & -             & -              \\
LDSGM \cite{wu2022ldsgm}                 & RoBERTa                     & 63.73              & 71.18              & 40.49                & 60.33               & 10.68              & 32.20               \\
ChatGPT \cite{chatgpt23} & - & 36.11 & 44.18 & 16.20 & 24.54 & - & - \\
\hdashline
GOLF (base)                  & RoBERTa                     & 65.76              & 72.52              & 41.74                & 61.16               & 11.79              & 32.85 \\
GOLF (large) & RoBERTa                     & \textbf{69.60}              & \textbf{74.67}              & \textbf{47.91}                & \textbf{63.91}               & \textbf{14.59}              & \textbf{42.35}
\\
\midrule
\midrule
\multicolumn{8}{c}{\textit{PDTB 3.0}}
\\\midrule
MANF \cite{xiang2022manf} & BERT & 56.63 & 64.04 & - & - & - & - \\
RoBERTa (Fine-tuning)               & RoBERTa                        & 68.31             & 71.59              & 50.63                & 60.14               & 14.72             & 39.43              \\
BMGF-RoBERTa \cite{liu2020bmgf}          & RoBERTa                     & 63.39              & 69.06              & -                & 58.13               & -             & -              \\
LDSGM \cite{wu2022ldsgm}                 & RoBERTa                     & 68.73              & 73.18              & 53.49                & 61.33               & 17.68              & 40.20               \\
ConnPrompt \cite{xiang2022connprompt} & RoBERTa & 69.51 & 73.84 & - & - & - & - \\
\hdashline
GOLF (base)                  & RoBERTa                     & 70.88              & 75.03              & 55.30                & 63.57               & 19.21              & 42.54 \\
GOLF (large) & RoBERTa                     & \textbf{74.21}              & \textbf{76.39}              & \textbf{60.11}                & \textbf{66.42}               & \textbf{20.66}              & \textbf{45.12}
\\
\bottomrule
\end{tabular}
\caption{\label{tab: main experiment}
Model comparison of multi-class classification on PDTB 2.0 and PDTB 3.0 in terms of macro-averaged F1 (\%) and accuracy (\%). }
\end{table*}

\subsection{Implementation Details}
We implement our model based on Huggingface’s transformers \cite{huggingface} and use the pre-trained RoBERTa \cite{liu2019roberta} (base or large version) as our Transformer encoder.
The layer number of MHIA and GCN are both set to 2. 
We set temperature $\tau$ in contrastive learning as 0.1.
We set $\Phi_1, \Phi_2, \Phi_3$ as a simple MLP with one hidden layer and \emph{tanh} activation function, which enables the gradient to be easily backpropagated to the encoder.
The node embeddings of senses with the dimension 100 are randomly initialized by \emph{kaiming\_normal} \cite{he2015delving}.
To avoid overfitting, we apply dropout with a rate of 0.1 after each GCN layer.
We adopt AdamW optimizer with a learning rate of 1e-5 and a batch size of 32 to update the model parameters for 15 epochs. 
The evaluation step is set to 100 and all hyperparameters are determined according to the best average model performance at three levels on the validation set. 
All experiments are performed five times with different random seeds and all reported results are averaged performance.

\subsection{Results}
\paragraph{Multi-label Classification Comparison}
The primary experimental results are presented in Table \ref{tab: main experiment}, which enables us to draw the following conclusions:
\begin{itemize}[]
\item Firstly, our GOLF model has achieved new state-of-the-art performance across all three levels, as evidenced by both macro-F1 and accuracy metrics. Specifically, on PDTB 2.0, GOLF (base) outperforms the current state-of-the-art LDSGM model \cite{wu2022ldsgm} by 2.03\%, 1.25\%, and 1.11\% in three levels, respectively, in terms of macro-F1. Additionally, it exhibits 1.34\%, 0.83\%, and 0.65\% improvements over the current best results in terms of accuracy. Moreover, in the case of PDTB 3.0, GOLF (base) also outperforms the current state-of-the-art ConnPrompt model \cite{xiang2022connprompt} by 1.37\% F1 and 1.19\% accuracy at the top level.
\item Secondly, employing RoBERTa-large embeddings in GOLF leads to a significant improvement in its performance. This observation indicates that our GOLF model can effectively benefit from larger pre-trained language models (PLMs).
\item Finally, despite the impressive performance of recent large language models (LLMs) such as ChatGPT \cite{openai2022chatgpt} in few-shot and zero-shot learning for various understanding and reasoning tasks \cite{DBLP:journals/corr/abs-2302-04023, DBLP:journals/corr/abs-2305-12870}, they still lag behind our GOLF (base) model by approximately 30\% in PDTB 2.0. This difference suggests that ChatGPT may struggle to comprehend the abstract sense of each discourse relation and extract the relevant language features from the text. Therefore, implicit discourse relation recognition remains a challenging and crucial task for the NLP community, which requires further exploration.
\end{itemize}

\begin{table}[]
\scriptsize
\centering
\begin{tabular}{lcccc}
\toprule
\textbf{Model}  & \makecell{\textbf{Exp.} \\ (53\%)} & \makecell{\textbf{Cont.} \\ (27\%)}          & \makecell{\textbf{Comp.} \\ (14\%)} & \makecell{\textbf{Temp.} \\ (3\%)}         \\
\midrule
BMGF \cite{liu2020bmgf}    & 77.66 & 60.98  & 59.44 & 50.26 \\
LDSGM \cite{wu2022ldsgm}   & 78.47 & 64.37  & 61.66 & 50.88 \\
\hdashline
GOLF (base) & 79.41 & 62.90  & 67.71 & 54.55 \\
GOLF (large) & \textbf{80.96} & \textbf{66.54}  & \textbf{69.47} & \textbf{61.40} \\
\bottomrule
\end{tabular}
\caption{\label{tab: top-level f1}
Label-wise F1 scores (\%) for the top-level senses of PDTB 2.0. The proportion of each sense is listed below its name.}
\end{table}

\begin{table}[]
\scriptsize
\centering
\begin{tabular}{lcccc}
\toprule
\textbf{Second-level Senses}  & \textbf{BMGF} & \textbf{LDSGM}          & \makecell{\textbf{GOLF} \\ (base)} & \makecell{\textbf{GOLF} \\ (large)}          \\
\midrule
Exp.Restatement (20\%)      & 53.83        & 58.06          & \textbf{59.84} & 59.03 \\
Exp.Conjunction (19\%)     & 60.17        & 57.91          & 60.28 & \textbf{61.54} \\
Exp.Instantiation (12\%)   & 67.96        & 72.60          & 75.36 & \textbf{77.98} \\
Exp.Alternative (1\%)     & 60.00        & 63.46          & \textbf{63.49} & 61.54 \\
Exp.List (1\%)            & 0.00          & 8.98           & 27.78 & \textbf{43.48} \\
\midrule
Cont.Cause (26\%)          & 59.60        & 64.36          & 65.35 & \textbf{65.98} \\
Cont.Pragmatic (1\%) & 0.00          & 0.00            & 0.00  & 0.00          \\
\midrule
Comp.Contrast (12\%)       & 59.75        & 63.52 & 61.95 & \textbf{67.57}         \\
Comp.Concession (2\%)    & 0.00          & 0.00            & 0.00 & \textbf{11.11}            \\
\midrule
Temp.Asynchronous (5\%)   & 56.18        & 56.47          & 63.82 & \textbf{65.49} \\
Temp.Synchrony (1\%)      & 0.00          & 0.00            & 0.00  & \textbf{13.33}   \\
\bottomrule
\end{tabular}
\caption{\label{tab: second-level f1}
Label-wise F1 scores (\%) for the second-level senses of PDTB 2.0. The proportion of each sense is listed behind its name.}
\end{table}

\paragraph{Label-wise Classification Comparison}
Here we present an evaluation of GOLF's performance on PDTB 2.0 using label-wise F1 comparison for top-level and second-level senses.
Table \ref{tab: top-level f1} showcases the label-wise F1 comparison for the top-level senses, demonstrating that GOLF significantly improves the performance of minority senses such as \textit{Temp} and \textit{Comp}. In Table \ref{tab: second-level f1}, we compare GOLF with the current state-of-the-art models for the second-level senses. Our results show that GOLF (base) enhances the F1 performance of most second-level senses, with a notable increase in \textit{Expa.List} from 8.98\% to 27.78\%. Furthermore, by using RoBERTa-large as embeddings, our GOLF (large) model breaks the bottleneck of previous work in two few-shot second-level senses, \textit{Temp.Synchrony} and \textit{Comp.Concession}. To further validate our model's ability of deriving better discourse relation representations, we compare the generated representations of GOLF with those of current state-of-the-art models for both top-level and second-level senses in Appendix \ref{sec: appendix c}.

\begin{table*}[htbp]
\small
\centering
\begin{tabular}{lcccccccc}
\toprule
\multirow{2}{*}{\textbf{Model}} & \multicolumn{2}{c}{\textbf{Top-level}} & \multicolumn{2}{c}{\textbf{Second-level}} & \multicolumn{2}{c}{\textbf{Connective}} & \multirow{2}{*}{\textbf{Top-Sec}} & \multirow{2}{*}{\textbf{Top-Sec-Conn}} \\
                       & $F_{1}$            & $Acc$           & $F_{1}$              & $Acc$            & $F_{1}$             & $Acc$           &                          &                               \\
\midrule
GOLF                   & \textbf{65.76}         & \textbf{72.52}         & \textbf{41.74 }          & \textbf{61.16}          & \textbf{11.79}          & \textbf{32.85}         & \textbf{59.65}                    & \textbf{27.55}                         \\
\hdashline
-\textit{w/o} MHIA   & 64.97              & 71.85              & 41.07                & 60.52               & 10.80               & 31.69              & 58.52                         & 26.18                              \\
-\textit{w/o} staircase          & 65.43              & 72.25              & 41.12                & 60.81               & 10.81               & 31.40              & 58.43                         & 26.08                              \\
-\textit{w/o} MHIA and staircase & 64.77 & 71.98 & 40.99 & 60.10 & 10.76 & 31.65 & 58.49 & 26.22
\\
-\textit{w/o} $\mathcal{L}_{G}$             & 65.37              & 71.61              & 40.78                & 60.40               & 11.56               & 32.73              & 59.01                         & 26.86                              \\
-\textit{w/o} $\mathcal{L}_{L}$              & 64.34              & 71.32              & 40.24                & 60.42               & 10.76               & 31.88              & 58.69                         & 26.37                              \\
-\textit{w/o} $\mathcal{L}_{G}$ and $\mathcal{L}_{L}$ & 63.85 & 71.04 & 39.98 & 59.92 & 10.72 & 30.47 & 58.23 & 25.89 \\
-\textit{r.p.} $\mathcal{L}_{L}$ with $\mathcal{L}_{L'}$           & 64.58              & 71.56 & 41.20               & 61.07                & 11.43              & 32.55    & 59.24 & 27.05                               \\
\bottomrule
\end{tabular}
\caption{\label{tab: ablation}
Ablation study on PDTB 2.0 considering the accuracy and F1 of each level as well as consistencies between hierarchies. “\textit{w/o}” stands for “without”; “\textit{r.p.}” stands for “replace”; "MHIA" stands for the Multi-Head Interactive Attention; $\mathcal{L}_{G}$ stands for the Global Hierarchy-aware Contrastive loss; $\mathcal{L}_{L}$ stands for the Local Hierarchy-aware Contrastive loss.}
\end{table*}

\begin{table}[htbp]
\small
\centering
\begin{tabular}{lcc}
\toprule
\textbf{Model}        & \textbf{Top-Sec}       &\textbf{ Top-Sec-Conn}   \\
\midrule
\midrule
\multicolumn{3}{c}{\textit{PDTB 2.0}} \\
\midrule
HierMTN-CRF & 46.29 & 19.15 \\
BMGF-RoBERTa & 47.06                       & 21.37          \\
LDSGM        & 58.61                       & 26.85          \\
\hdashline
GOLF (base)  & 59.65   & 27.55 \\
GOLF (large) & \textbf{61.79} & \textbf{36.00} \\
\midrule
\midrule
\multicolumn{3}{c}{\textit{PDTB 3.0}} \\
\midrule
HierMTN-CRF & 50.19   & 27.82 \\
BMGF-RoBERTa & 52.33   & 29.16          \\
LDSGM        & 60.32   & 34.57          \\
\hdashline
GOLF (base)  & 61.31   & 36.97 \\
GOLF (large) & \textbf{64.86} & \textbf{38.26} \\
\bottomrule
\end{tabular}
\caption{\label{tab: consistency}
Comparison with current state-of-the-art models on the consistency among multi-level sense predictions.}
\end{table}

\paragraph{Multi-level Consistency Comparison}
Following \cite{wu2022ldsgm}, we evaluate the consistency among multi-level sense predictions via two metrics: 1) Top-Sec: the percentage of correct predictions at both the top-level and second-level senses; 2) Top-Sec-Con: the percentage of correct predictions across all three level senses.
Our model's results, as displayed in Table \ref{tab: consistency}, demonstrate more consistent predictions than existing state-of-the-art models in both Top-Sec and Top-Sec-Con, verifying the effectiveness of our model in integrating global and local hierarchical information.

\section{Ablation Study}
\label{sec:analyses}
Firstly, we investigate the efficacy of individual modules in our framework.
For this purpose, we remove the Multi-Head Interactive Attention (MHIA), the "staircase" in Classifier, the Global Hierarchy-aware Contrastive loss $\mathcal{L}_{G}$, and the Local Hierarchy-aware Contrastive loss $\mathcal{L}_{L}$ from GOLF one by one.
Note that removing the "staircase" in Classifier means that we keep the cross-entropy loss but remove the dependence between logits from different hierarchical levels.
Table \ref{tab: ablation} indicates that eliminating any of the four modules would hurt the performance across all three levels and reduce the consistency among multi-level label predictions. 
At the same time, the Local Hierarchy-aware Contrastive loss contributes mostly.
Besides, removing both the Global Hierarchy-aware Contrastive loss $\mathcal{L}_{G}$ and the Local Hierarchy-aware Contrastive loss $\mathcal{L}_{L}$ significantly hurts the performance.
The results show that incorporating label hierarchies from both the global and local perspectives is indeed beneficial.
Secondly, we replace the Local Hierarchy-aware Contrastive loss $\mathcal{L}_{L}$ (Equation (\ref{equation: soft local})) with the hard-label version $\mathcal{L}_{L'}$ (Equation (\ref{equation: hard local})) and find that the performance drops notably.
It verifies the usefulness of the scoring function in Equation \ref{equation: score}, which considers more subtle semantic structures of local hierarchy.
In Appendix \ref{sec: appendix b}, We also analyze the effects of various hyperparameters consisting of the number layer of MHIA and GCN, the coefficients $\lambda_1$ and $\lambda_2$, and the temperature $\tau$.

\section{Conclusion}
In this paper, we present a novel Global and Local Hierarchy-aware Contrastive Framework for implicit discourse relation recognition (IDRR). 
It can sufficiently incorporate global and local hierarchies to learn better discourse relation representations with the aid of multi-task learning and contrastive learning.
Compared with current state-of-the-art approaches, our model empirically reaches better performance at all hierarchical levels of the PDTB dataset and achieves more consistent predictions on multi-level senses.

\section*{Limitations}
In this section, we illustrate the limitations of our method, which could be summarized into the following two aspects.


Firstly, since the cumbersome data annotation leads to few publicly available datasets of IDRR tasks, we only conduct experiments on English corpora including PDTB 2.0 and PDTB 3.0. 
In the future, we plan to comprehensively evaluate our model on more datasets and datasets in other languages.

Secondly, considering that instances of PDTB are contained in paragraphs of the Wall Street Journal articles, our approach ignores wider paragraph-level contexts beyond the two discourse arguments.
As shown in \cite{dai2018pdrr}, positioning discourse arguments in their wider context of a paragraph may further benefit implicit discourse relation recognition.
It is worth exploring how to effectively build wider-context-informed discourse relation representations and capture the overall discourse structure from the paragraph level.

\section*{Ethics Statement}
Since our method relies on pre-trained language models, it may run the danger of inheriting and propagating some of the models' negative biases from the data they have been pre-trained on \cite{BenderGMS21bias}. Furthermore, we do not see any other potential risks.

\section*{Acknowledgments}

W.\ Wang was supported by HKUST(GZ) Grant G0101000028, GZU-HKUST Joint Research Collaboration Grant GZU22EG04, and Guangzhou Municipal Science and Technology Project (No. 2023A03J0003). 

\newpage
\bibliography{anthology,custom}
\bibliographystyle{acl_natbib}

\clearpage

\appendix

\section{Data Statistics}
\label{sec: appendix a}

\begin{table}[ht]
\small
\centering
\begin{tabular}{lccc}
\toprule
\textbf{Second-level Senses}  & \textbf{Train}  & \textbf{Dev}   & \textbf{Test}  \\
\midrule
Exp.Conjunction      & 2,814  & 258   & 200   \\
Exp.Restatement      & 2,430  & 260   & 211   \\
Exp.Instantiation    & 1,100  & 106   & 118   \\
Exp.List             & 330    & 9     & 12    \\
Exp.Alternative      & 150    & 10    & 9     \\
\midrule
Cont.Cause           & 3,234  & 281   & 269   \\
Cont.Pragmatic cause & 51     & 6     & 7     \\
\midrule
Comp.Contrast        & 1,569  & 166   & 128   \\
Comp.Concession     & 181    & 15    & 17    \\
\midrule
Temp.Asynchronous    & 540    & 46    & 54    \\
Temp.Synchrony       & 148    & 8     & 14    \\
\midrule
Total                & 12,547 & 1,165 & 1,039 \\
\bottomrule
\end{tabular}
\caption{\label{tab: second-level senses pdtb2}
The data statistics of second-level senses in PDTB 2.0.}
\end{table}

\begin{table}[ht]
\small
\centering
\begin{tabular}{lccc}
\toprule
\textbf{Second-level Senses}  & \textbf{Train}  & \textbf{Dev}   & \textbf{Test}  \\
\midrule
Exp.Conjunction      & 3,566  & 298   & 237   \\
Exp.Level-of-detail      & 2,698  & 274   & 214   \\
Exp.Instantiation    & 1,215  & 117   & 127   \\
Exp.Manner      & 1,159    & 57    & 53     \\
Exp.Substitution    & 405    & 32     & 31    \\
Exp.Equivalence      & 256  & 25   & 30   \\
\midrule
Cont.Cause           & 4,280  & 423   & 388   \\
Cont.Purpose & 688     & 66     & 59     \\
Cont.Cause+Belief  & 140  & 13   & 14   \\
Cont.Condition &138 &17 & 14 \\
\midrule
Comp.Concession     & 1,159    & 105    & 97    \\
Comp.Contrast        & 813  & 87   & 62   \\
\midrule
Temp.Asynchronous    & 1,025    & 103    & 105    \\
Temp.Synchronous       & 331    & 24     & 35    \\
\midrule
Total                & 17,873 & 1,641 & 1,466 \\
\bottomrule
\end{tabular}
\caption{\label{tab: second-level senses pdtb3}
The data statistics of second-level senses in PDTB 3.0.}
\end{table}

\section{Visualization of Discourse Relation Representations}
\label{sec: appendix c}

\begin{figure*}[!t]
	\centering 
	\includegraphics[width=\linewidth]{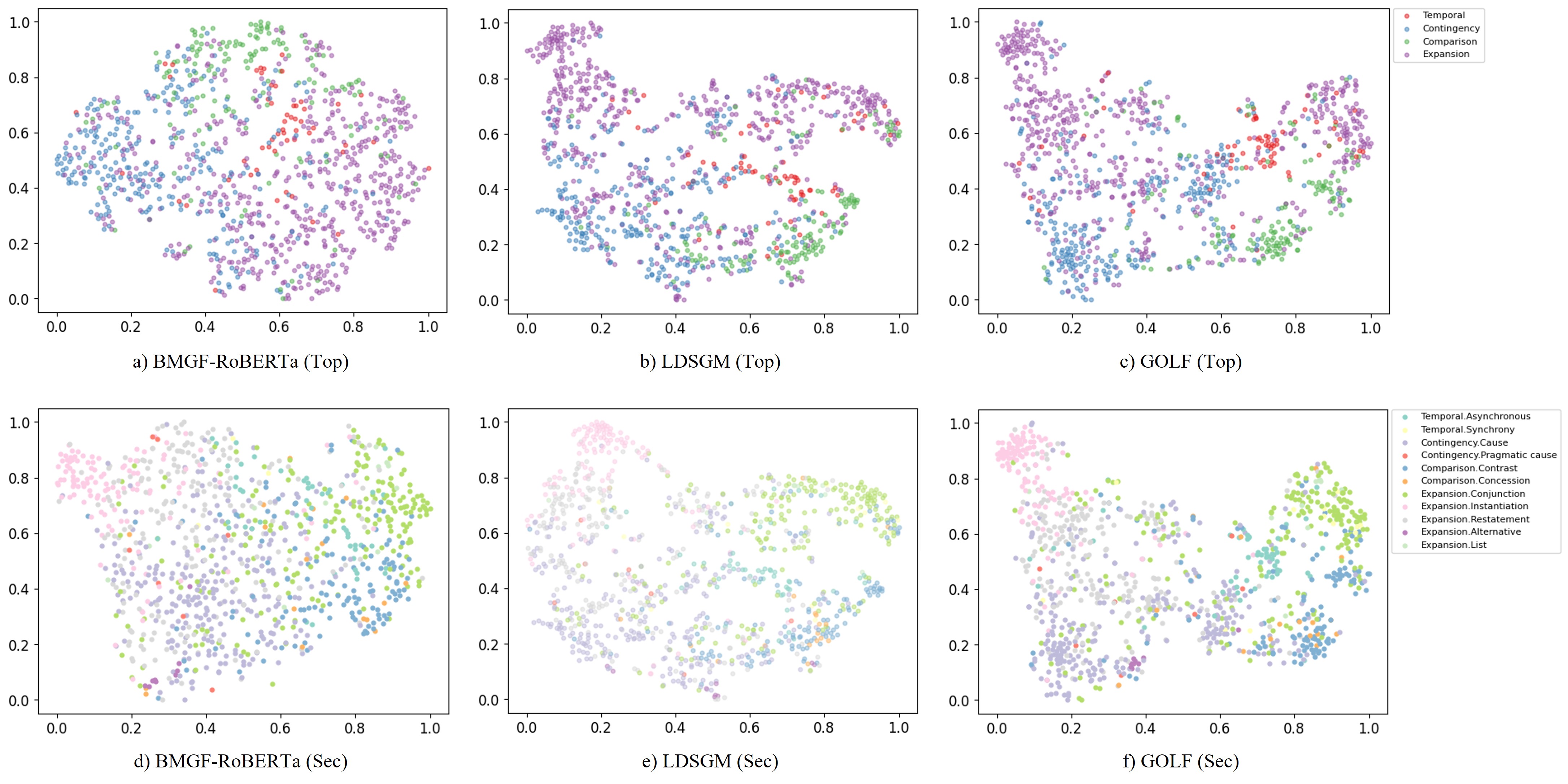}
	\caption{t-SNE visualization of discourse relation representations for the top-level and second-level senses on PDTB 2.0 test set.}
	\label{fig: distribution}
\end{figure*}

Here we investigate the quality of discourse relation representations generated by our GOLF model with visualization aids.
Figure \ref{fig: distribution} depicts the 2D t-SNE \cite{van2008visualizing} visualization of discourse relation representations for top-level and second-level senses on the PDTB 2.0 test set.
As we can see, compared with current state-of-the-art models BMGF-RoBERTa \cite{liu2020bmgf} and LDSGM \cite{wu2022ldsgm}, our model can generate more centralized discourse relation representations belonging to the same senses (\textit{e.g.}, \textit{Temporal} at the top level, marked in red), and more separated representations belonging to different senses.
It verifies our model's capability of deriving better discourse relation representations.

\section{Effects of Hyperparameters}
\label{sec: appendix b}
\begin{figure*}[!bp]
	\centering 
	\includegraphics[width=\linewidth]{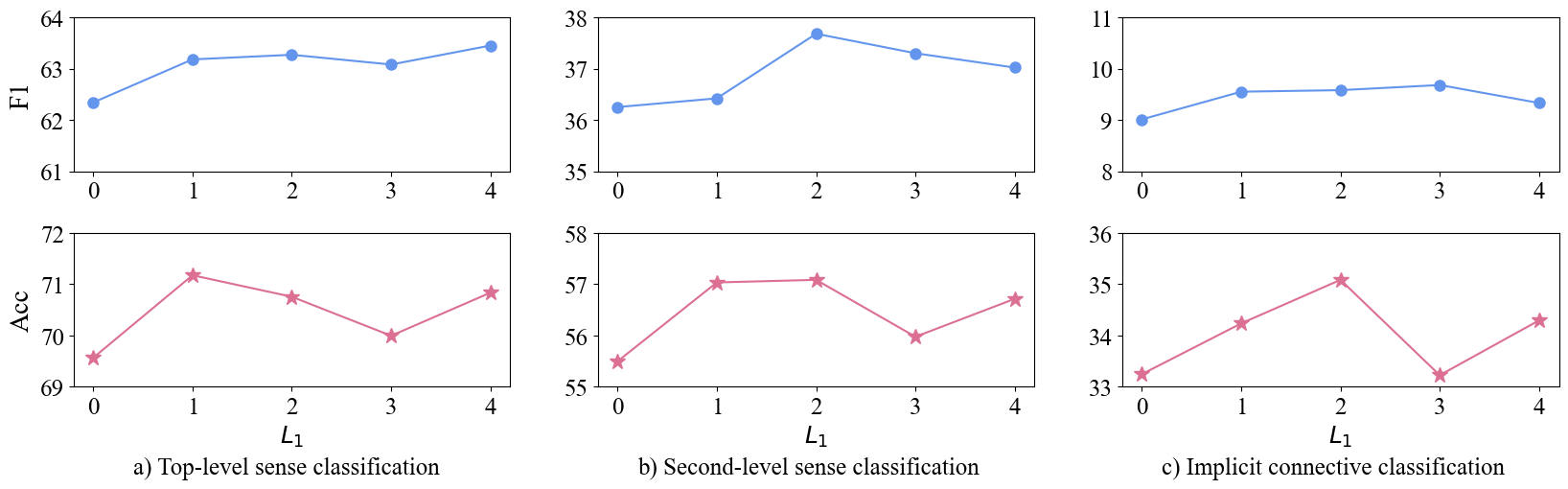}
	\caption{Effects of the number layer $L_1$ of MHIA on the development set.}
	\label{fig: l_1}
\end{figure*}

\begin{figure*}[!bp]
	\centering 
	\includegraphics[width=\linewidth]{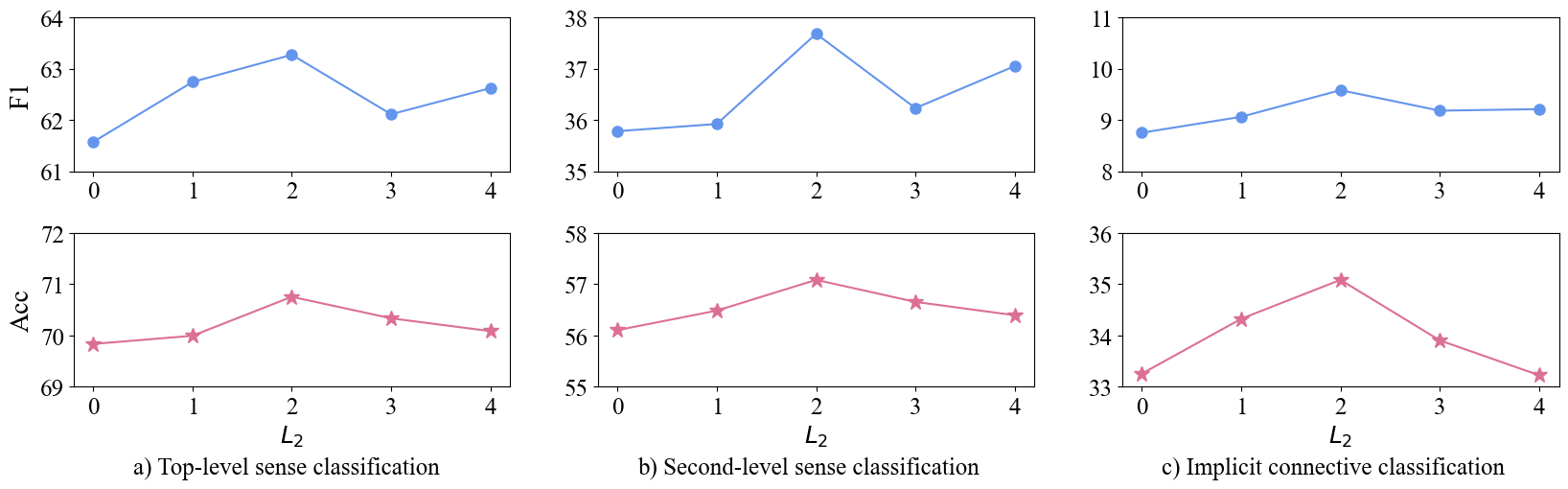}
	\caption{Effects of the number layer $L_2$ of GCN on the development set.}
	\label{fig: l_2}
\end{figure*}

\begin{figure*}[!bp]
	\centering 
	\includegraphics[width=\linewidth]{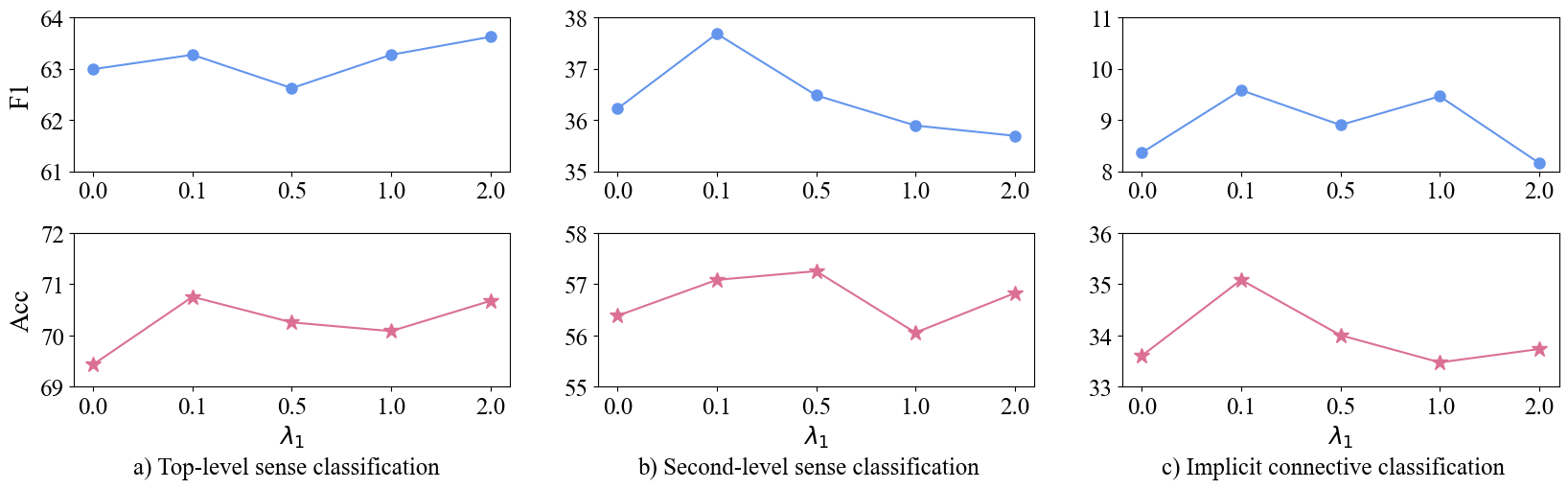}
	\caption{Effects of the coefficient $\lambda_1$ of the global hierarchy-aware contrastive loss on the development set.}
	\label{fig: lambda_1}
\end{figure*}

\begin{figure*}[!bp]
	\centering 
	\includegraphics[width=\linewidth]{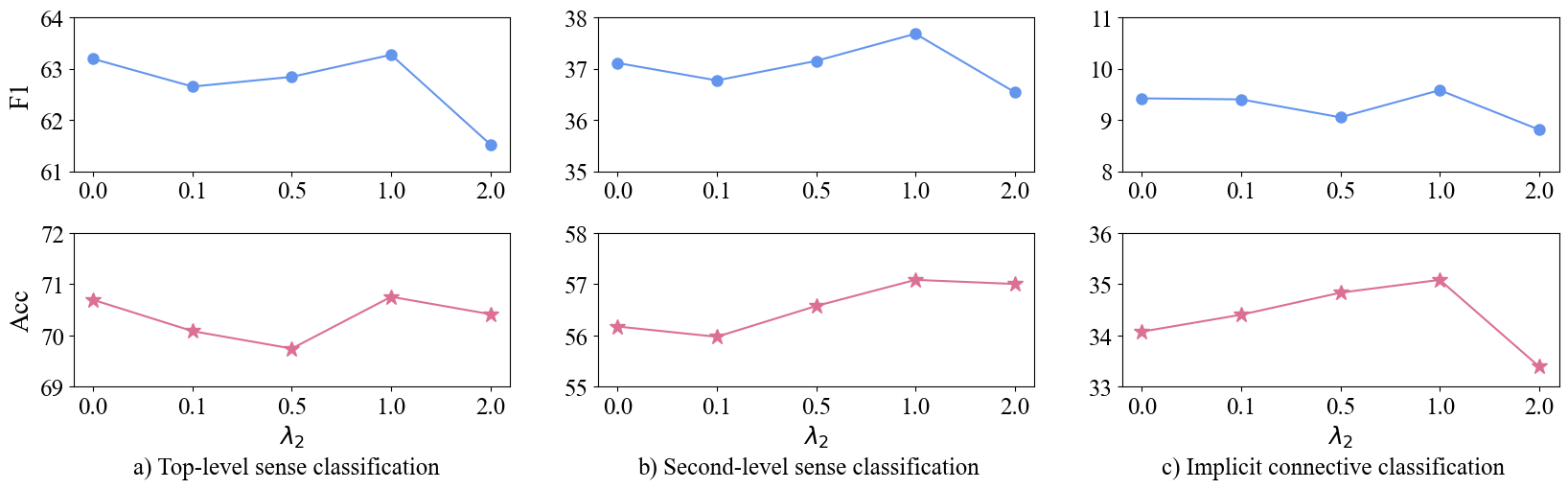}
	\caption{Effects of the coefficient $\lambda_2$ of the local hierarchy-aware contrastive loss on the development set.}
	\label{fig: lambda_2}
\end{figure*}

\begin{figure*}[!bp]
	\centering 
	\includegraphics[width=\linewidth]{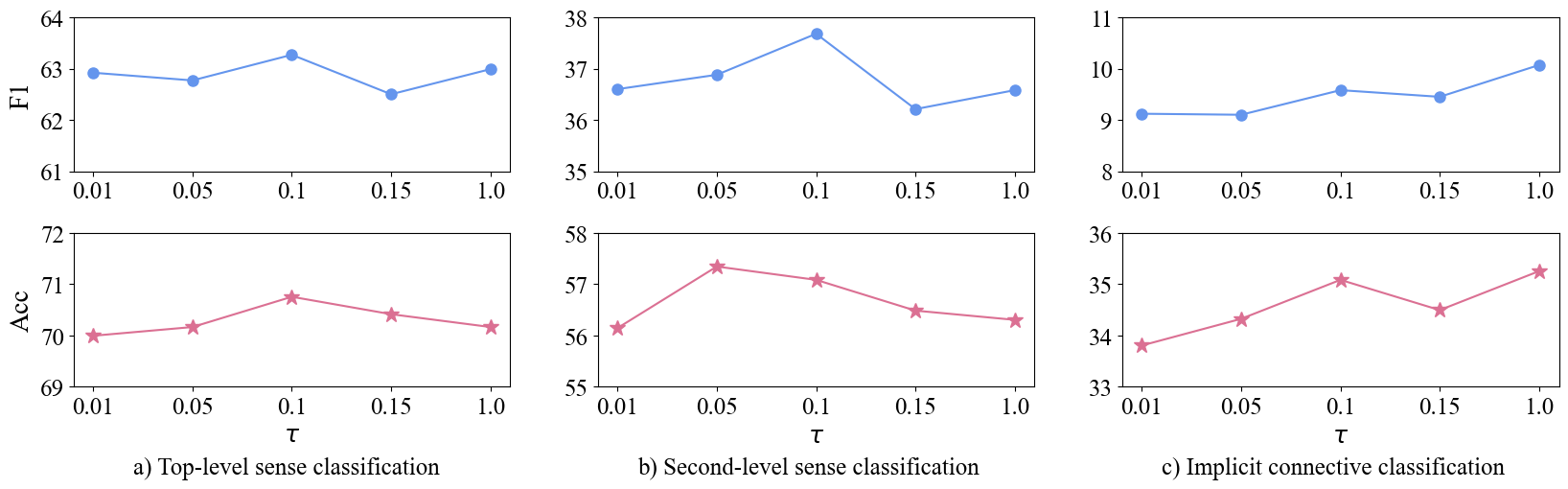}
	\caption{Effects of the temperature $\tau$ in contrastive learning on the development set.}
	\label{fig: temp}
\end{figure*}

Here we investigate the effects of various hyperparameters on the development set of PDTB 2.0. 
These hyperparameters include the number layer $L_1$ of MHIA (Figure \ref{fig: l_1}), the number layer $L_2$ of GCN (Figure \ref{fig: l_2}), the coefficient $\lambda_1$ of the global hierarchy-aware contrastive loss (Figure \ref{fig: lambda_1}),  the coefficient $\lambda_2$ of the local hierarchy-aware contrastive loss (Figure \ref{fig: lambda_2}), and the temperature $\tau$ in contrastive learning (Figure \ref{fig: temp}).
Note that we only change one hyperparameter at a time.

\section{Label-wise Classification on PDTB 3.0}
\label{sec: appendix d}

\begin{table}[h]
\small
\centering
\begin{tabular}{lcc}
\toprule
\textbf{Top-level Senses}  & \textbf{GOLF} (base) & \textbf{GOLF} (large)          \\
\midrule
Exp (47\%) & 80.01 & \textbf{80.50} \\
Cont (32\%) & 74.54 & \textbf{74.83}      \\
Comp (11\%) & 64.67 & \textbf{71.59}     \\
Temp (10\%) & 64.80 & \textbf{70.92} \\
\bottomrule
\end{tabular}
\caption{\label{tab: top-level f1 pdtb3}
Label-wise F1 scores (\%) for the top-level senses of PDTB 3.0. The proportion of each sense is listed behind its name.}
\end{table}

\begin{table}[h]
\small
\centering
\begin{tabular}{lcc}
\toprule
\textbf{Second-level Senses}  & \makecell{\textbf{GOLF} \\ (base)} & \makecell{\textbf{GOLF} \\ (large)}          \\
\midrule
Exp.Conjunction (16\%) & \textbf{64.09} & 63.69 \\
Exp.Level-of-detail (15\%)  & 52.60 & \textbf{59.29} \\
Exp.Instantiation (9\%) & 72.53 & \textbf{73.77} \\
Exp.Manner (4\%) & \textbf{63.53} & 62.61 \\
Exp.Substitution (2\%) & 66.67 & \textbf{72.22} \\
Exp.Equivalence (2\%) & \textbf{25.39} & 24.00 \\
\midrule
Cont.Cause (26\%) & 69.47 & \textbf{72.49} \\
Cont.Purpose (4\%) & 71.60 & \textbf{72.73} \\
Cont.Cause+Belief (1\%) & 0.00 & 0.00 \\
Cont.Condition (1\%) & 66.67  & \textbf{92.31}          \\
\midrule
Comp.Concession (7\%) & 59.09 & \textbf{63.37}      \\
Comp.Contrast (4\%) & 43.33 & \textbf{60.27}     \\
\midrule
Temp.Asynchronous (7\%) & 68.79 & \textbf{77.55} \\
Temp.Synchronous (2\%) & 41.00  & \textbf{42.27}   \\
\bottomrule
\end{tabular}
\caption{\label{tab: second-level f1 pdtb3}
Label-wise F1 scores (\%) for the second-level senses of PDTB 3.0. The proportion of each sense is listed behind its name.}
\end{table}

\end{document}